\def\paperID{13008}
\def\confName{CVPR}
\def\confYear{2024}

\def\paperTitle{Fast and Efficient: Mask Neural Fields for 3D Scene Segmentation}

\def\authorBlock{
    Zihan Gao \qquad
    Lingling Li \thanks{Corresponding author.}\qquad
    Licheng Jiao \qquad
    Fang Liu \\
    Xu Liu \qquad 
    Wenping Ma \qquad 
    Yuwei Guo  \qquad
    Shuyuan Yang \\
    School of Artificial Intelligence, Xidian University \\
    {\tt\small z1han\_gao@163.com \qquad llli@xidian.edu.cn}
}

\newif\ifreview 
\newif\ifarxiv \newcommand{\arxiv}{\arxivtrue}
\newif\ifcamera 
\newif\ifrebuttal 

\arxiv 
\pdfoutput=1
\documentclass[10pt,twocolumn,letterpaper]{article}
\ifreview \usepackage[review]{cvpr} \fi
\ifarxiv \usepackage[pagenumbers]{cvpr} \fi
\ifrebuttal \usepackage[rebuttal]{cvpr} \fi
\ifcamera \usepackage{cvpr} \fi

\usepackage{graphicx}	
\usepackage{amsmath}	
\usepackage{amssymb}	
\usepackage{booktabs}
\usepackage{times}
\usepackage{microtype}
\usepackage{epsfig}
\usepackage[table,xcdraw,dvipsnames]{xcolor}
\usepackage{caption}
\usepackage{float}
\usepackage{placeins}
\usepackage{color, colortbl}
\usepackage{stfloats}
\usepackage{enumitem}
\usepackage{tabularx}
\usepackage{xstring}
\usepackage{multirow}
\usepackage{xspace}
\usepackage{url}
\usepackage{subcaption}
\usepackage{xcolor}
\usepackage[hang,flushmargin]{footmisc}

\ifcamera \usepackage[accsupp]{axessibility} \fi

% ...

\newcommand{\nbf}[1]{{\noindent \textbf{#1.}}}

\ifarxiv  \fi

\newcommand{\R}[1]{{%
    \textbf{%
        \ifstrequal{#1}{1}{\textcolor{red}{R#1}}{%
        \ifstrequal{#1}{2}{\textcolor{blue}{R#1}}{%
        \ifstrequal{#1}{3}{\textcolor{magenta}{R#1}}{%
        \ifstrequal{#1}{4}{\textcolor{teal}{R#1}}{%
                           \textcolor{cyan}{R#1}%
        }}}}%
    }%
}}

\usepackage{xr-hyper}

\makeatletter
\newcommand*{\addFileDependency}[1]{
  \typeout{(#1)}
  \@addtofilelist{#1}
  \IfFileExists{#1}{}{\typeout{No file #1.}}
}

\makeatother
\newcommand*{\myexternaldocument}[1]{
    \externaldocument{#1}
    \addFileDependency{#1.tex}
    \addFileDependency{#1.aux}
}

\definecolor{cvprblue}{rgb}{0.21,0.49,0.74}
\usepackage[pagebackref,breaklinks,colorlinks,citecolor=cvprblue]{hyperref}
\usepackage[capitalize]{cleveref}
\crefname{section}{Sec.}{Secs.}
\crefname{table}{Table}{Tables}
\crefname{figure}{Fig.}{Figs.}

\ifarxiv \crefname{appendix}{App.}{Apps.}
\else \crefname{appendix}{Suppl.}{Suppls.} \fi

\unless\ifarxiv \myexternaldocument{_supplementary} \fi

\begin{document}
\setlength{\belowcaptionskip}{-0.4cm}

\title{\paperTitle}
\author{\authorBlock}
\maketitle

\begin{abstract}
Understanding 3D scenes is a crucial challenge in computer vision research with applications spanning multiple domains.  Recent advancements in distilling 2D vision-language foundation models into neural fields, like NeRF and 3DGS, enable open-vocabulary segmentation of 3D scenes from 2D multi-view images without the need for precise 3D annotations. However, while effective, these methods typically rely on the per-pixel distillation of high-dimensional CLIP features, introducing ambiguity and necessitating complex regularization strategies, which adds inefficiency during training. This paper presents MaskField, which enables efficient 3D open-vocabulary segmentation with neural fields from a novel perspective. Unlike previous methods, MaskField decomposes the distillation of mask and semantic features from foundation models by formulating a mask feature field and queries. MaskField overcomes ambiguous object boundaries by naturally introducing SAM segmented object shapes without extra regularization during training. By circumventing the direct handling of dense high-dimensional CLIP features during training, MaskField is particularly compatible with explicit scene representations like 3DGS. Our extensive experiments show that MaskField not only surpasses prior state-of-the-art methods but also achieves remarkably fast convergence. We hope that MaskField will inspire further exploration into how neural fields can be trained to comprehend 3D scenes from 2D models. Please find the code at \url{https://github.com/keloee/MaskField}
\end{abstract}
\section{Introduction}
\label{sec:intro}

Recent developments in 3D scene representation, notably Neural Radiance Fields (NeRF) ~\cite{mildenhall2020nerf} and 3D Gaussian Splatting (3DGS) ~\cite{kerbl20233d}, have paved the way for learning-based methods to recover 3D geometry from merely posed 2D images. These advances have spurred research into leveraging 2D foundation models, contextualized by neural fields, to comprehend 3D scenes from multi-view 2D images \cite{lerf2023,liu2023weakly,qin2023langsplat, zhou2023feature}. Utilizing pre-trained models like CLIP ~\cite{radford2021learning}, these methods enable 3D open-vocabulary semantic segmentation without actual 3D data and annotation. It holds significant potential in diverse applications such as robot navigation ~\cite{shafiullah2022clip}, autonomous driving ~\cite{yang2023unisim, yang2023emernerf}, and urban planning ~\cite{hu2020towards,zhang2022artificial}. 

\begin{figure*}
    \centering
    \includegraphics[width=1\linewidth]{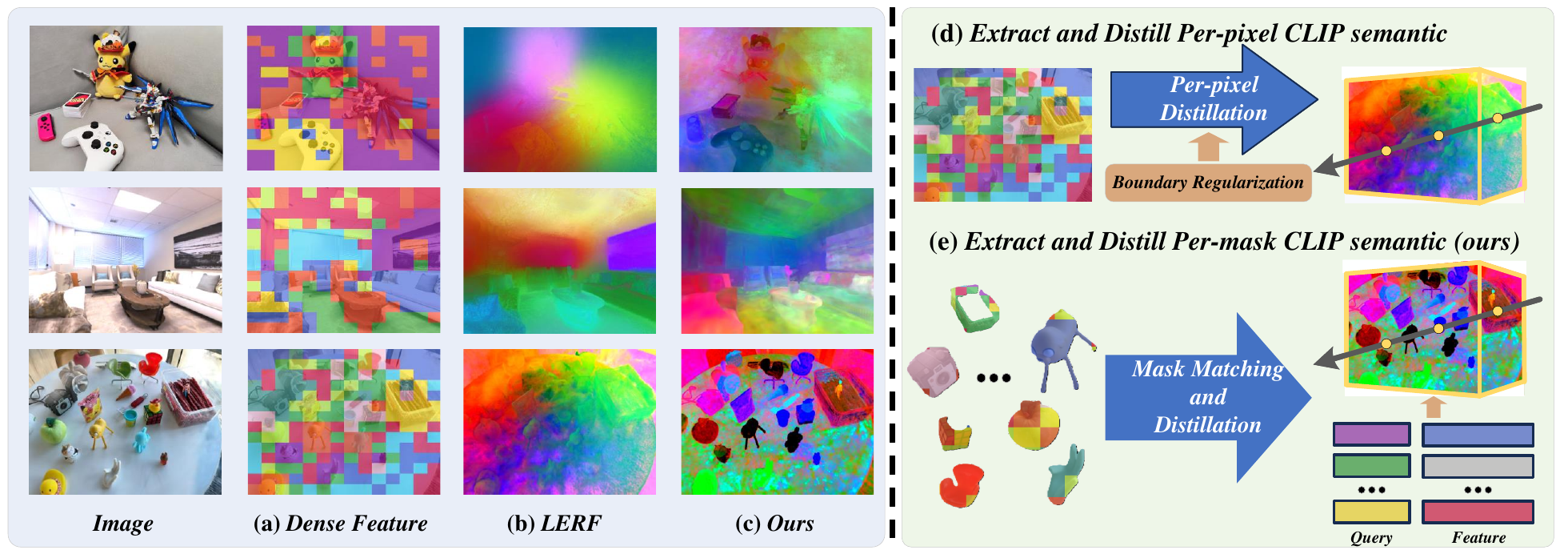}
    \caption{\textbf{(a)} Extracting pixel-aligned CLIP features from image crops shows ambiguity around object boundary. \textbf{(b)} PCA visualization of LERF CLIP feature field. \textbf{(c)} Our method produces clear object boundary  \textbf{(d)} Previous method extracts dense CLIP features and performs distillation at pixel-level. \textbf{(e)} Our proposed MaskField performs mask-level distillation to avoid handling the pixel-aligned high-dimensional ambiguous CLIP feature during training.}
    \label{fig:intro}
\vspace{-0.2cm}
\end{figure*}

However, since CLIP is trained at the image level, extracting pixel-aligned features for per-pixel segmentation introduces ambiguity. To illustrate this, we visualize feature maps generated by various methods in \cref{fig:intro}. Addressing this ambiguity, prior approaches to 3D open-vocabulary semantic segmentation with neural fields have employed complex frameworks. Typically, they first construct a multi-scale feature pyramid by extracting CLIP embeddings from image crops at various scales for each view. These multi-scale features are then distilled into a neural feature field, with additional refinement of object boundaries using off-the-shelf models such as DINO \cite{caron2021emerging} or SAM \cite{kirillov2023segment}. While effective, this approach is highly inefficient, requiring extensive resources to manage dense, high-dimensional CLIP features and complex regularization during training. For example, \cite{liu2023weakly} trains a small scene with 28 views can demand up to 64GB of RAM and 0.5 hours just to load preprocessed CLIP features, plus an additional 1.3 hours of training on an RTX 3090 GPU. Such inefficiencies limit these methods’ real-world applicability. This raises a key question: \textit{Can we bypass per-pixel CLIP feature distillation while still achieving accurate segmentation?}

In this paper, we adopt a novel perspective and present MaskField, a novel approach inspired by prior research \cite{cheng2021per, ding2022decoupling} to overcome the inefficiencies of per-pixel CLIP feature distillation. Instead of performing per-pixel distillation, MaskField decouples the distillation of shape and semantics and performs distillation at the mask level. Specifically, MaskField utilizes a neural field to represent mask features in 3D. These features are rendered using techniques tailored to neural fields, ensuring multi-view consistency. Additionally, we introduce a set of scene-level learnable queries, each paired with a high-dimensional feature token. These tokens, designed to be invariant to 3D position or viewing direction, enable consistent predictions across different views. Each query corresponds to a specific mask in the scene: the query aids in binary mask prediction through a dot product operation with the rendered mask features, while the associated feature token holds distilled CLIP features, allowing it to differentiate the semantic content of each mask. The predicted masks are supervised using SAM-generated masks with a bipartite matching-based assignment strategy. Our contribution can be summarized as follows:
\begin{enumerate}
\item We propose MaskField, a novel paradigm for 3D scene segmentation that leverages foundation models and neural fields. By performing mask-level distillation and integrating SAM’s object boundaries, MaskField eliminates the need for additional boundary regularization, enhancing segmentation quality and reducing training time.
\item By decoupling mask and semantic feature distillation, MaskField supports low-dimensional mask features in the neural feature field, offering compatibility with 3D Gaussian Splatting (3DGS) and lowering computational demands for complex scenes.
\item We evaluate MaskField on three benchmark datasets, demonstrating its effectiveness, robustness, and efficiency across diverse 3D segmentation tasks.
\end{enumerate}
\section{Related Work}
\label{sec:related}
\subsection{Neural Feature Fields.}
Researchers have developed learning-based methods to represent 3D scenes, primarily categorized into Neural Radiance Field (NeRF) \cite{mildenhall2020nerf,barron2021mip,barron2022mip} and 3D Gaussian Splatting (3DGS) \cite{kerbl20233d,yu2023mip}. NeRF uses a multi-layer perceptron (MLP) to implicitly represent scenes by modeling 3D coordinates and viewing directions, producing corresponding RGB and volume density values. Conversely, 3DGS employs explicit 3D Gaussians for scene representation. Both methods can be adapted to train a feature field that generates high-dimensional feature vectors by distilling pre-trained vision models, enhancing 3D semantic understanding.

Distilled Feature Fields \cite{kobayashi2022decomposing} initially explored distilling pixel-aligned feature vectors from LSeg \cite{li2021language}, but this approach struggles with generalization in long-tail distribution classes. LERF \cite{lerf2023} proposed extracting a multi-scale feature pyramid from image crops to distill CLIP features for open-vocabulary segmentation. Extending this idea, 3D-OVS \cite{liu2023weakly} introduced multi-scale and multi-spatial strategies to adapt CLIP’s image-level features for pixel-level segmentation, adding two regularization terms to mitigate ambiguities in CLIP features. Building on the success of 3DGS in novel view synthesis, Langsplat \cite{qin2023langsplat} and Feature GS \cite{zhou2023feature} uses 3DGS to create 3D representations. Langsplat also advocat for the use of the hierarchy defined by the Segment Anything Model (SAM) to address CLIP feature ambiguities.

Although these methods show impressive results by performing pixel-level distillation of CLIP features into a neural feature field, constructing pixel-aligned CLIP features is inherently challenging due to their ambiguities. In this work, we perform mask-level distillation by leveraging neural fields as scene-level mask generators to circumvent directly addressing these ambiguities. 

\subsection{Open-vocabulary Segmentation.}
In recent years, open-vocabulary segmentation has seen considerable growth, fueled by the widespread availability of extensive text-image datasets and advanced pre-trained models like CLIP \cite{radford2021learning}. Researchers have approached problem from different perspectives. A direct method \cite{li2021language} leverages the capabilities of CLIP to align pixel-level visual features with its text embeddings. More recent studies \cite{ghiasi2022scaling,ding2022decoupling,xu2022simple,liang2023open,xu2023open} have focused on using class-agnostic mask generators that classify masks via a parallel CLIP branch. Another emerging approach \cite{zhou2023zegclip,xu2023side} incorporates learnable tokens or adapter layers to predict masks directly using a frozen CLIP. However, these methods, trained at the image level, lead to inconsistent segmentation across multiple views of a 3D scene. Building on advancements in 2D open-vocabulary segmentation, MaskField utilizes neural fields as a scene-level mask generator, harnessing their inherent multi-view consistency and transferring the open-vocabulary capabilities of CLIP into 3D environments.

\subsection{Foundation Models}
Pre-trained foundation models \cite{caron2021emerging,radford2021learning,cherti2023reproducible,kirillov2023segment} have become a cornerstone in the field of computer vision. These models are foundational for developing a comprehensive understanding of visual content, trained on vast datasets with a high number of parameters. For instance, CLIP \cite{radford2021learning,cherti2023reproducible} combines an image encoder with a text encoder, utilizing an image-text contrastive learning strategy to form associations between images and their corresponding text descriptions from large-scale image-text data \cite{schuhmann2022laionb}. Demonstrating significant zero-shot capabilities \cite{cherti2023reproducible}, CLIP also integrates well with other modules, enhancing various downstream tasks \cite{li2021language,ramesh2021zero}. Beyond CLIP, the Segment Anything Model (SAM) \cite{kirillov2023segment} serves as a foundation model specifically for image segmentation. SAM is trained on an extensive dataset of 11 million images and 1 billion masks, enabling it to generate high-quality, class-agnostic region proposals that perform robustly across different applications \cite{yu2023inpaint,zhang2023personalize,ke2024segment,ma2024segment}. While CLIP and SAM show impressive capabilities in 2D image understanding, training such a model in 3D is challenging due to the expensive 3D annotations and unstructured scene representation. In this work, we distill the capabilities of SAM and CLIP into 3D to enable efficient 3D open-vocabulary semantic segmentation.

\section{Method}
\label{sec:method}
\begin{figure*}
    \centering
    \includegraphics[width=1\linewidth]{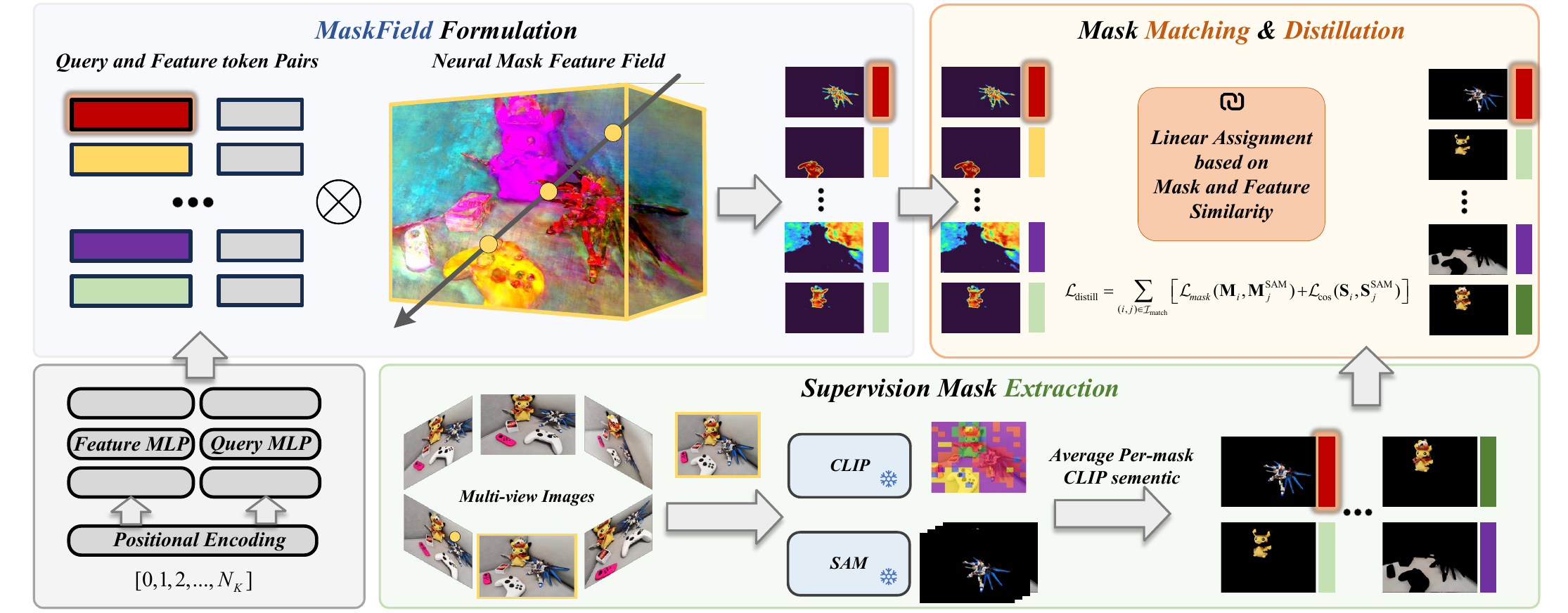}
    \caption{\textbf{An overview of the proposed MaskField.} Given a set of multi-view images, our method distills the open-vocabulary knowledge from CLIP at a mask level. Our method naturally introduces region boundaries segmented by SAM without the need for complex regularization during training.}
    \label{fig:method}
\vspace{-0.2cm}
\end{figure*}
Given a set of multi-view 2D images, our objective is to reconstruct a neural field that is semantic aware. Previously, this problem has been addressed by associating each 3D point with a CLIP feature to represent its semantic meaning \cite{lerf2023, liu2023weakly}. As CLIP only generates image-level features, extracting pixel-aligned CLIP features presents challenges. Departing from these approaches, we introduce a method to avoid the inefficiencies associated with the inherent ambiguities of CLIP features by segmenting the neural field at a mask level. An overview of MaskField is presented in \cref{fig:method}.

In this section, we first revisit the challenges of modeling neural fields for 3D open-vocabulary semantic segmentation and highlight the key factors contributing to inaccuracy and inefficiency. We then elaborate on the proposed MaskField, demonstrating how it effectively addresses these challenges. 

\subsection{Revisiting Neural Fields for 3D Open-vocabulary Semantic Segmentation} 
Given a set of \(N\) calibrated multi-view images, a neural field is trained to represent the 3D scene. This exploration aims to understand the inherent inefficiencies in current learning paradigms. The specific type of neural field is not crucial for this discussion. For ease of reference, we denote the neural field as \(\Phi(\mathbf{x}, \mathbf{d}) \rightarrow (\mathbf{c}, \sigma)\), where \(\mathbf{x}\) represents the 3D coordinates, \(\mathbf{d}\) the viewing direction, \(\mathbf{c}\) the color, and \(\sigma\) the density. For a pixel \(u\), the color can be rendered in 2D as follows:
\begin{equation}
\mathbf{\hat{C}}(u)=\sum_i T_i \alpha_i \mathbf{c}_i, \\
T_i=\Pi_{j=0}^{i-1}(1-\alpha_i), 
\alpha_i=1-\exp(-\delta_i\sigma_i).
\end{equation}
Here, \(T_i\) represents the accumulated transmittance, and \(\alpha_i\) denotes the opacity of the point. Building on this, most existing methods \cite{kobayashi2022decomposing, lerf2023, liu2023weakly, qin2023langsplat} construct a neural feature field, denoted as \(\Psi_{f}(\mathbf{x}, \mathbf{d}) \rightarrow (\mathbf{c}, \sigma, \mathbf{f})\), which not only models the scene’s geometry but also integrates high-dimensional semantic features vector \(\mathbf{f}\). The feature map can be rendered in 2D similarly to the color:
\begin{equation}
\begin{aligned}
\mathbf{\hat{F}}(u)=\sum_i T_i \alpha_i \mathbf{f}_i.
\label{eq:2}
\end{aligned}
\end{equation}
This enriched representation enables the feature distillation of the CLIP \cite{radford2021learning} image encoder based on the geometry provided by the neural field. However, the fundamental design of CLIP, which focuses on generating image-level features \(F_{V}\in \mathbb{R}^{D_{S}}\), introduces a challenge when precise neural feature field distillation requires pixel-aligned features \(\mathbf{F} \in \mathbb{R}^{D_{S} \times H \times W}\). Also, the high-dimensional CLIP feature introduces large memory usage and slow rasterization with explicit neural fields such as 3DGS.

As illustrated in \cref{fig:intro}, employing CLIP for pixel-aligned distillation inevitably confronts the issue of feature ambiguity. To bridge this gap between image and pixel-aligned requirements, mainstream methods segment the image into patches \cite{lerf2023, liu2023weakly} , or smaller regions \cite{qin2023langsplat} . This approach transforms the broad, image-wide CLIP features into a more localized form that can be directly applied at the pixel level. Also, recent methods introduce autoencoders \cite{qin2023langsplat} or upsamplers \cite{zhou2023feature} to reduce the feature dimension.

Although pixel-level distillation of CLIP features into a neural field has shown effectiveness, there are two main factors contributing to its inaccuracy and inefficiency. \textbf{First}, the inherent ambiguity in CLIP features complicates achieving precise segmentation. This necessitates complex regularization strategies to clarify object boundaries, which in turn increases the computational load during training. \textbf{Second}, the per-pixel distillation of high-dimensional, pixel-aligned CLIP features often leads to inefficiency, placing significant demands on system resources. This limitation restricts scalability and practical application in larger scenes.

Due to the inaccuracy and inefficiency associated with the per-pixel distillation of CLIP features, we explore a mask-level strategy that avoids distilling pixel-aligned CLIP features and demonstrates to be more effective and efficient.

\subsection{MaskField Formulation}

We introduce MaskField, a novel approach for 3D open-vocabulary semantic segmentation with neural fields. Unlike prior methods that distill CLIP features on a per-pixel basis, MaskField separates shape and semantic distillation, operating at the mask level. This approach generates and classifies object masks, effectively mitigating the complexities of high-dimensional CLIP feature ambiguity.

We formulate MaskField as a mask neural field \(\Psi_{m}(\mathbf{x}, \mathbf{d}) \rightarrow (\mathbf{c}, \sigma, \mathbf{f}_{m})\), where \(\mathbf{f}_{m}\) is a mask feature vector of dimension \(D_{m}\). Using \(N_{K}\) query and feature token pairs \(\left\{(\mathbf{Q}_i, \mathbf{S}_i) \mid i=1, \dots, N_K\right\}\), where \(\mathbf{Q}_i \in \mathbb{R}^{D_{m}}\) and \(\mathbf{S}_i \in \mathbb{R}^{D_{S}}\), MaskField renders a mask feature map \(\mathbf{\hat{F}}_{m} \in \mathbb{R}^{D_{m} \times H \times W}\) from a given viewpoint. Binary masks \(\mathbf{M}_{i}\) are then derived as \(\mathbf{M}_{i} = \mathbf{\hat{F}}_{m} \cdot \mathbf{Q}_{i}\), producing masks in \(\mathbb{R}^{H \times W}\) for each object. Each feature token \(\mathbf{S}_{i}\) corresponds to the CLIP embedding for \(\mathbf{M}_{i}\). In practice, the query and feature tokens are generated through two shallow MLPs, with input as the Fourier features \cite{tancik2020fourier} of their corresponding IDs. Notably, both query and semantic tokens remain viewpoint-independent. Leveraging the multi-view consistency of neural fields, these tokens serve as scene-level mask representations, ensuring consistent segmentation across different viewpoints.

\noindent \textbf{Training.}
To establish supervision for MaskField, we extract class-agnostic masks, denoted as \(\left\{\textbf{M}^{\text{SAM}}_{j} \in \mathbb{R}^{H \times W} \mid j=1,2, \ldots, N\right\}\), by inputting a regular grid of \(32 \times 32\) point prompts into SAM. To associate these masks with semantics, we use the method from \cite{zhou2022extract, fu2024featup} to extract visual CLIP features \(\mathbf{F}_{\text{visual}}\). Although this extraction typically yields coarse features \cite{wu2023clipself, fu2024featup}, which previous methods \cite{liu2023weakly, lerf2023} have found challenging for precise localization, MaskField’s mask-level distillation mitigates dependence on high localization accuracy.

To train the parameters of MaskField, we need to match the predicted masks \(\textbf{M}\) with \(\mathbf{M^{\text{SAM}}}\). Similar to previous works in query-based segmentation \cite{cheng2021per, wang2021max, ghiasi2022scaling}, MaskField adopts a bipartite matching-based assignment strategy\footnote{the matching cost is computed the same as \(\mathcal{L}_{\text{distill}}\) in \cref{eq:distill}} to align each predicted mask with a mask predicted by SAM. Given the set of matched indices \(\mathcal{I}_{\text{match}} = \{(i, j) \mid \textbf{M}{i} \text{ matches } \textbf{M}_{j}^{\text{SAM}}\}\), the loss function is defined as a weighted sum of two components: a mask distillation loss and a feature distillation loss. The mask distillation loss consists of a dice loss \cite{milletari2016v} and focal loss \cite{lin2017focal} to refine the binary mask, while the feature distillation loss uses cosine similarity to align feature representations.
\begin{equation}
\begin{aligned}
&\mathcal{L}_{\text{mask}} = \sum_{(i, j) \in \mathcal{I}_{\text{match}}} \left[\mathcal{L}_{\text{focal}}(\mathbf{M}_{i}, \mathbf{M}_{j}^{\text{SAM}}) + \lambda \mathcal{L}_{\text{dice}}(\mathbf{M}_{i}, \mathbf{M}_{j}^{\text{SAM}})\right], \\
&\mathcal{L}_{\text{feature}} =  \sum_{(i, j) \in \mathcal{I}_{\text{match}}} \mathcal{L}_{\cos}(\mathbf{S}_{i}, \mathbf{S}_{j}^{\text{SAM}}), \\
&\mathcal{L}_{\text{distill}} =\mathcal{L}_{\text{mask}} + \mathcal{L}_{\text{feature}}.    
\end{aligned}
\label{eq:distill}
\end{equation}
The extra remaining unmatched masks are penalized to zero by optimizing the following loss functions:
\begin{equation}
\begin{aligned}
\mathcal{L}_{\text{extra}} = \sum_{i \notin \mathcal{I}_{\text{match}}}\left \| \mathbf{M}_{i}  \right \| ^2_2.
\label{eq:5}
\end{aligned}
\end{equation}

\nbf{Inference} For inference, we address two tasks: open-vocabulary segmentation and open-vocabulary query. The former segments the scene based on a predefined list of text labels, while the latter segments object within the scene based on a specific text query.

In open-vocabulary segmentation, given class descriptions for \(C\) classes, the CLIP text encoder generates text features \(\mathbf{F}_{T}\in \mathbb{R}^{D_S}\) . The relevance is computed as the cosine similarity between the distilled CLIP feature and the text features of each class: \(p = \cos\left(\mathbf{F}_{\text{text}}, \textbf{S}\right)\). Instead of using the absolute relevance value for segmentation, we normalize the relevance across masks. To produce the final segmentation map, we first apply non-maximum suppression to eliminate overlapping masks and retain those with high relevance scores, effectively selecting masks with appropriate scales. We then aggregate the remaining masks via matrix multiplication, determining the label of each pixel by:
\begin{equation}
\begin{aligned}
l(u) = \arg\max \sum_{i} p \cdot \mathbf{M}_{i}(u),
\label{eq:6}
\end{aligned}
\end{equation}
where masks segmented by SAM are effectively combined based on their classification scores and mask confidence.

For open-vocabulary query, we calculate the relevance score of each mask to a given text query, following the approach in LERF \cite{lerf2023}, and filter out masks with relevance scores below a specified threshold. Additionally, masks with low activation (penalized by \(\mathcal{L}_{\text{extra}}\)) are removed. The remaining masks are then composed to form the final object segmentation. More details are provided in the appendix.

MaskField simplifies segmentation by eliminating the need for pixel-aligned, high-dimensional CLIP features, instead averaging CLIP semantics on a per-mask basis and performing distillation at the mask level. By separating mask and semantic distillation, MaskField allows for low-dimensional shape features \(\mathbf{f}_{m}\) in the neural feature field, reducing computational overhead and streamlining the workflow. MaskField also naturally incorporates SAM’s object boundaries without additional regularization, enabling efficient training with low-resolution settings suitable for NeRF and 3DGS. As a result, MaskField achieves state-of-the-art performance with only 5 minutes of training, representing a 19.5x speed-up over prior methods ~\cite{liu2023weakly}.

\section{Experiment}
\label{sec:experiment}
\begin{table}[t]
\definecolor{red}{rgb}{1,0.6,0.6}
\definecolor{orange}{rgb}{1,0.8,0.6}
\definecolor{yellow}{rgb}{1,1,0.6}
  \centering
  \caption{\textbf{Quantitative comparisons on mIoU and mBIoU on the LERF-Mask dataset.} We report the mIoU($\uparrow$) and mBIoU($\uparrow$) scores of the following methods and highlight the \colorbox{red}{best}, \colorbox{orange}{second-best}, and \colorbox{yellow}{third-best} scores.}
  \resizebox{0.95\linewidth}{!}{
  \begin{tabular}{c|cc|cc|cc}
    \toprule
    \toprule
    Scene & \multicolumn{2}{c|}{figurines} & \multicolumn{2}{c|}{ramen} & \multicolumn{2}{c}{teatime} \\
    \midrule
    Metric & \textbf{mIoU} & \textbf{mBIoU} & \textbf{mIoU} & \textbf{mBIoU} & \textbf{mIoU} & \textbf{mBIoU} \\
    \midrule
    DEVA \cite{cheng2023tracking} & 46.2 & 45.1  & 56.8  & 51.1  & 54.3  & 52.2  \\
    LERF \cite{lerf2023} & 33.5  & 30.6  & 28.3  & 14.7  & 49.7  & 42.6  \\
    SA3D \cite{cen2023segment}  & 24.9  & 23.8  & 7.4  & 7.0  & 42.5  & 39.2  \\
    Langsplat \cite{qin2023langsplat}   & 52.8  & 50.5  & 50.4  & 44.7  & 69.5  & 65.6  \\
    Gaussian Grouping \cite{ye2023gaussian}   & \cellcolor{yellow}69.7  & \cellcolor{yellow}67.9  & \cellcolor{yellow}77.0  & \cellcolor{yellow}68.7  &  \cellcolor{orange}71.7  &  \cellcolor{orange}66.1  \\
    \midrule
    MaskField(NeRF) & \cellcolor{red}79.5  & \cellcolor{red}75.8  & \cellcolor{orange}85.3  & \cellcolor{orange}77.8  & \cellcolor{yellow}68.2  & \cellcolor{yellow}65.3  \\
    MaskField(3DGS) & \cellcolor{orange}74.2  & \cellcolor{orange}71.6  & \cellcolor{red}86.9  & \cellcolor{red}80.4  &  \cellcolor{red}72.8  &  \cellcolor{red}70.1  \\
    \bottomrule
    \bottomrule
    \end{tabular}%
    }
  \label{tab:lerf}%
  \vspace{-0.5cm}
\end{table}%
\begin{figure*}
    \centering
    \includegraphics[width=0.95\linewidth]{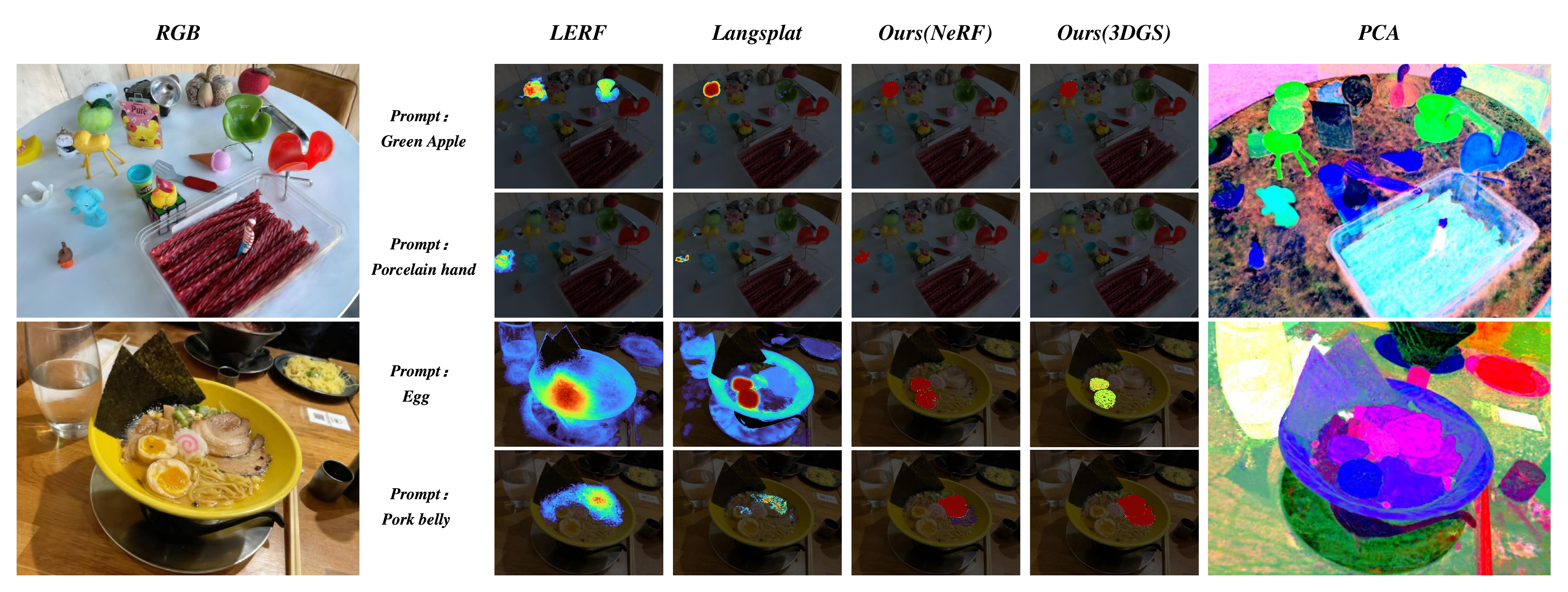}
    \caption{\textbf{Qualitative comparisons of 2 different scenes in LERF-Mask dataset.} Our method successfully gives the most accurate object segmentation.}
    \label{fig:lerf}
\end{figure*}
\begin{figure*}
    \centering
    \includegraphics[width=0.95\linewidth]{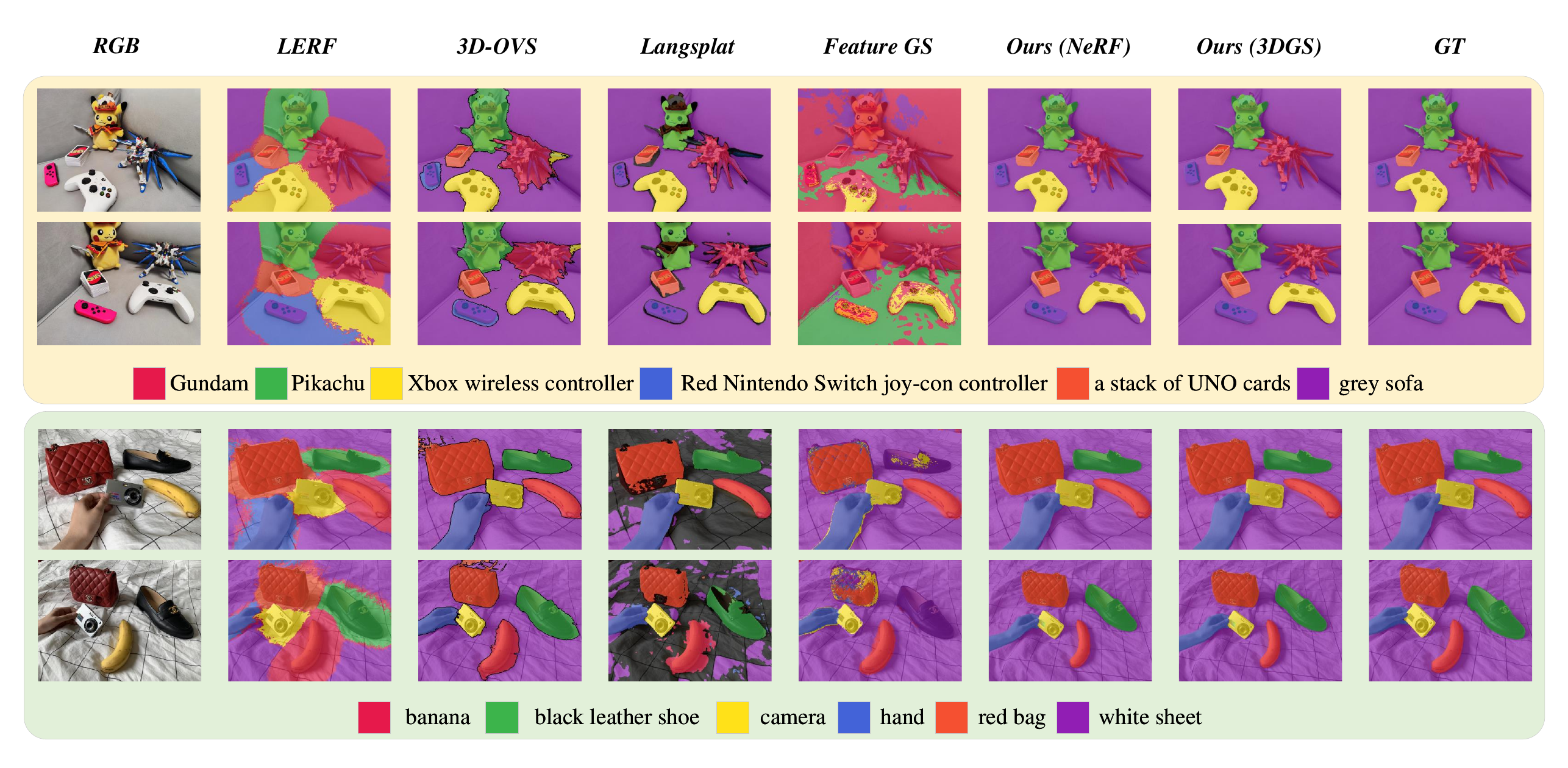}
    \caption{\textbf{Qualitative comparisons of 2 different scenes in 3DOVS dataset.} Our method successfully recognizes long-tailed objects and gives the most accurate segmentation maps.}
    \label{fig:3dovs}
\end{figure*}
\begin{table*}[t]
    \definecolor{red}{rgb}{1,0.6,0.6}
    \definecolor{orange}{rgb}{1,0.8,0.6}
    \definecolor{yellow}{rgb}{1,1,0.6}
    \centering
    \caption{
    \textbf{Quantitative comparisons on mIoU and Acc on the 3D-OVS datatset.} We report the mIoU($\uparrow$) and Acc($\uparrow$) scores of the following methods and highlight the \colorbox{red}{best}, \colorbox{orange}{second-best}, and \colorbox{yellow}{third-best} scores.
    }
    \label{tab:comparision_iou}
    \resizebox{0.95\linewidth}{!}{
    \begin{tabular}{@{}l@{\,\,}|@{\,\,}c@{\,\,}|cc|cc|cc|cc|cc|cc|cc|cc|cc|cc@{}}
    \midrule
    \midrule
    & \multicolumn{1}{c|}{\multirow{2}{*}{Methods}}  & \multicolumn{2}{c|}{\multirow{2}{*}{\textit{bed}}} & \multicolumn{2}{c|}{\multirow{2}{*}{\textit{sofa}}} & \multicolumn{2}{c|}{\multirow{2}{*}{\textit{lawn}}} & \multicolumn{2}{c|}{\multirow{2}{*}{\textit{room}}} & \multicolumn{2}{c|}{\multirow{2}{*}{\textit{bench}}} & \multicolumn{2}{c|}{\multirow{2}{*}{\textit{table}}} & \multicolumn{2}{c|}{\multirow{1}{*}{\textit{blue}}} & \multicolumn{2}{c|}{\multirow{1}{*}{\textit{covered}}} & \multicolumn{2}{c|}{\multirow{2}{*}{\textit{snacks}}} & \multicolumn{2}{c}{\multirow{1}{*}{\textit{office}}} \\
    & & & & & & & & & & & & & & \multicolumn{2}{c|}{\textit{sofa}} & \multicolumn{2}{c|}{\textit{desk}} & & & \multicolumn{2}{c}{\textit{desk}} \\
    
    & & \small \textbf{mIoU} & \small \textbf{Acc} & \small \textbf{mIoU} & \small \textbf{Acc} & \small \textbf{mIoU} & \small \textbf{Acc} & \small \textbf{mIoU} & \small \textbf{Acc} & \small \textbf{mIoU} & \small \textbf{Acc} & \small \textbf{mIoU} & \small \textbf{Acc} & \small \textbf{mIoU} & \small \textbf{Acc} & \small \textbf{mIoU} & \small \textbf{Acc} & \small \textbf{mIoU} & \small \textbf{Acc} & \small \textbf{mIoU} & \small \textbf{Acc}\\

    \midrule
    \midrule
    \multirow{2}{*}{2D} & LSeg~\cite{li2021language} &  56.0 & 87.6 &  04.5 & 16.5 & 17.5 & 77.5 &  19.2 & 46.1 & 06.0 & 42.7 &  07.6 & 29.9 & 17.6 & 68.9 & 21.8 & 79.4 & 32.1 & 84.5 & 19.6 & 41.5 \\
& ODISE~\cite{xu2023open} &  52.6	&86.5		&48.3		&35.4		&39.8		&82.5	&52.5	&59.7	&24.1		&39.0	&39.7	&34.5	&55.4	&56.1	&67.3	&75.0	&33.7	&63.4	&55.5	&61.0\\
\midrule
\multirow{4}{*}{NeRF} & FFD~\cite{kobayashi2022decomposing} &  56.6	&86.9	&03.7	&09.5	&42.9	&82.6	&25.1	&51.4	&06.1	&42.8	&07.9	&30.1	&16.3	&77.6	&26.8	&82.6	&34.9	&84.1	&19.0	&41.6\\
& Sem(ODISE)~\cite{zhi2021place} &  50.3	&86.5	&27.7	&22.2	&24.2	&80.5	&29.5	&61.5	&25.6	&56.4	&18.4	&30.8	&52.8	&58.1	&45.8	&61.4	&23.4	&57.0	&65.4	&88.3 \\
& LERF~\cite{lerf2023} &  73.5	&86.9	&27.0	&43.8	&73.7	&93.5	&46.6	&79.8	&53.2	&79.7	&33.4	&41.0	&32.2	&85.0	&51.8	&91.4	&48.7	&91.5	&43.9	&88.2\\
& 3D-OVS~\cite{liu2023weakly} & 89.5		&96.7		&74.0	&91.6	&88.2	&97.3	&92.8	&98.9	&\cellcolor{yellow}89.3	&\cellcolor{yellow}96.3	&\cellcolor{yellow}88.8	&\cellcolor{yellow}96.5	&82.8	&97.7	&\cellcolor{yellow}88.6	&\cellcolor{yellow}97.2	&\cellcolor{orange}95.8	&\cellcolor{orange}99.1	&91.7	&\cellcolor{orange}96.2\\
\midrule
\multirow{3}{*}{3DGS}
& Langsplat~\cite{qin2023langsplat} &   73.5		&89.7	&\cellcolor{yellow}82.3 &\cellcolor{red}98.7	&89.9	&95.6	&\cellcolor{red}95.0	&\cellcolor{red}99.4	&70.6	&92.6	&77.8	&92.3	&\cellcolor{yellow}94.5	&\cellcolor{yellow}99.4	&80.2	&92.1	&92.0	&96.3	&\cellcolor{red}93.6	&\cellcolor{red}99.1 \\
& Feature GS~\cite{zhou2023feature} &   56.6		&87.5		&06.7		&12.4		&37.3	&82.6	&20.5	&36.7	&06.2	&43.0	&08.3	&32.2	&18.0	&77.3	&23.8	&80.6	&32.5	&84.7	&20.0	&41.8 \\
& Gaussian Grouping~\cite{ye2023gaussian} & \cellcolor{yellow}95.7	&\cellcolor{yellow}98.1	&74.9	&	97.1	&\cellcolor{yellow}90.4	&	\cellcolor{yellow}98.6	&63.5	&	90.6	&	82.3	&	96.1	&70.1	&	92.2	&69.4	&	92.3	&61.7	&	89.2	&81.3	&	86.2	&75.3	&	96.1\\
\midrule
& \textbf{Ours (NeRF)} & \cellcolor{orange}96.7	& \cellcolor{orange}99.2 &   \cellcolor{orange}93.8 & \cellcolor{yellow}98.3 &   \cellcolor{orange}92.9 & \cellcolor{orange}98.8 &   \cellcolor{orange}92.9 & \cellcolor{orange}97.9 & \cellcolor{red}94.4 & \cellcolor{red}98.7 & \cellcolor{red}94.3 & \cellcolor{red}98.5 & \cellcolor{orange}95.0 & \cellcolor{orange}99.5 & \cellcolor{orange}90.1 & \cellcolor{orange}97.7 & \cellcolor{red}95.8 &\cellcolor{yellow}99.2 & \cellcolor{red}91.7&\cellcolor{orange}96.2\\
& \textbf{Ours (3DGS)} & \cellcolor{red}97.6& \cellcolor{red}99.4 &\cellcolor{red}94.6 & \cellcolor{red}98.7 &   \cellcolor{red}94.4 & \cellcolor{red}99.0 &  \cellcolor{yellow}92.4 & \cellcolor{yellow}98.3 &  \cellcolor{orange}92.8 & \cellcolor{orange}98.0 & \cellcolor{orange}93.6 & \cellcolor{orange}98.1 & \cellcolor{red}96.0 & \cellcolor{red}99.6 & \cellcolor{red}90.8 & \cellcolor{red}97.9 & \cellcolor{yellow}92.7 & \cellcolor{red}98.0 & \cellcolor{orange}92.1 & \cellcolor{orange}96.2 \\
    \midrule
    \midrule
    \end{tabular}
    }
\end{table*}
\begin{table}[t]
\definecolor{red}{rgb}{1,0.6,0.6}
\definecolor{orange}{rgb}{1,0.8,0.6}
\definecolor{yellow}{rgb}{1,1,0.6}
  \centering
  \caption{\textbf{Quantitative comparisons on mIoU and Acc on the Replica dataset.} We report the mIoU($\uparrow$) and Acc($\uparrow$) scores of the following methods and highlight the \colorbox{red}{best}, \colorbox{orange}{second-best}, and \colorbox{yellow}{third-best} scores.}
  \resizebox{0.95\linewidth}{!}{
  \begin{tabular}{c|cc|cc|cc}
    \toprule
    \toprule
    Scene & \multicolumn{2}{c|}{room0} & \multicolumn{2}{c|}{room1} & \multicolumn{2}{c}{office3} \\
    \midrule
    Metric & \textbf{mIoU} & \textbf{Acc} & \textbf{mIoU} & \textbf{Acc} & \textbf{mIoU} & \textbf{Acc} \\
    \midrule
    3D-OVS \cite{liu2023weakly} & 21.7  & 39.1 & 32.5  & 44.6  & 26.7  & 36.6  \\
    Feature GS\cite{zhou2023feature} & 42.7  & 59.0  & \cellcolor{yellow}64.8  & \cellcolor{orange}76.5  & 40.7  & 61.3  \\
    FFD\cite{kobayashi2022decomposing}   & 43.3  & 61.0  & 63.5  & 75.1  & 38.3  & \cellcolor{yellow}61.7  \\
    LERF\cite{lerf2023}   & \cellcolor{yellow}45.7  & \cellcolor{yellow}61.1  & 59.0  & 70.5  & \cellcolor{yellow}45.5  & 60.6  \\
    \midrule
    MaskField(NeRF) & \cellcolor{red}71.9  & \cellcolor{red}80.7  & \cellcolor{red}71.8  & \cellcolor{red}76.6  & \cellcolor{red}51.6  & \cellcolor{red}73.7  \\
    MaskField(3DGS) & \cellcolor{orange}68.9  & \cellcolor{orange}77.7  & \cellcolor{orange}66.1  & \cellcolor{yellow}74.2  & \cellcolor{orange}47.9  & \cellcolor{orange}61.9  \\
    \bottomrule
    \bottomrule
    \end{tabular}%
    }
  \label{tab:replica}%
  \vspace{-0.6cm}
\end{table}%
\begin{figure*}
    \centering
    \includegraphics[width=0.95\linewidth]{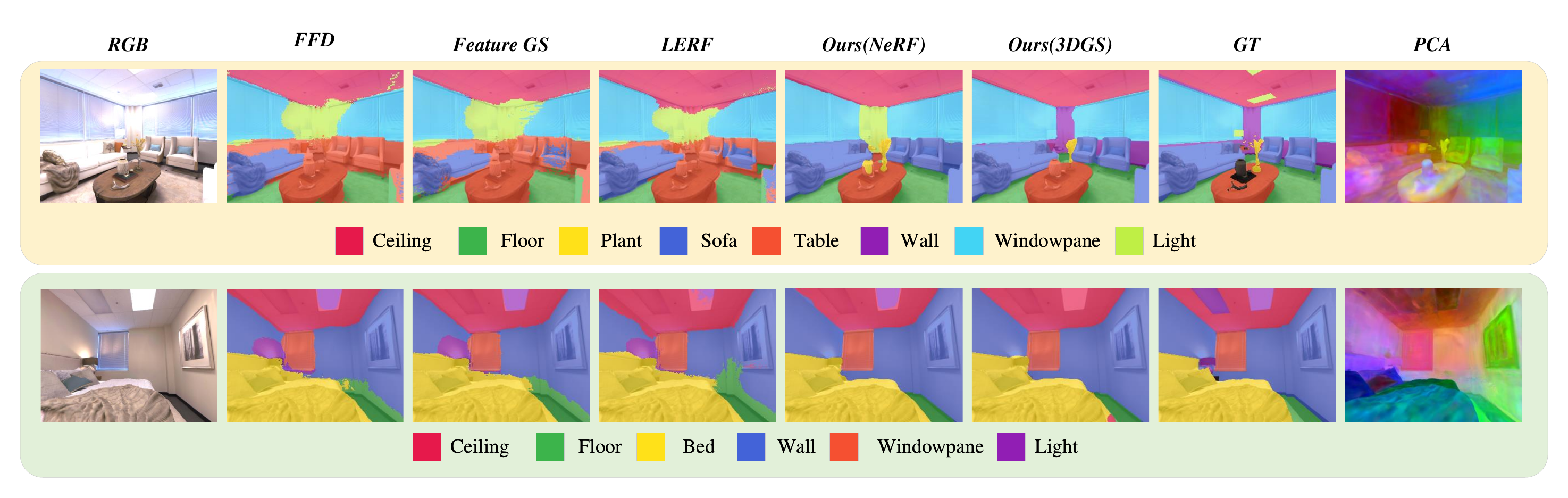}
    \caption{\textbf{Qualitative comparisons of 2 different scenes in Replica dataset.} Our method successfully recognizes objects in complex geometry and gives the most accurate segmentation maps.}
    \label{fig:replica}
\end{figure*}
\subsection{Implementation Details.}
We implement MaskField using PyTorch, and all experiments are conducted on a GeForce RTX 3090 GPU with 32GB RAM. The CLIP features are extracted using MaskCLIP ~\cite{zhou2022extract} and FeatUP ~\cite{fu2024featup}. Additionally, we implement MaskField with 3DGS ~\cite{kerbl20233d} following the setting of Feature GS  ~\cite{zhou2023feature}. Unless otherwise 
specified, the mask neural field feature dimension is set to 16 for both NeRF and 3DGS. For each scene, we first train a neural field using RGB for geometry. Then, we train our mask neural field by fixing all other parameters during the reconstruction stage. We provide more details on implementation and explain the choice of dataset and comparison method in the appendix.

\subsection{Open-vocabulary Query.}
For open-vocabulary querying, we evaluate MaskField on the LERF-Mask benchmark \cite{ye2023gaussian}, with results shown in \cref{tab:lerf} and \cref{fig:lerf}. MaskField exhibits a clear advantage over prior state-of-the-art approaches, significantly outperforming them in handling the inherent ambiguity of CLIP features. Unlike LERF-based similarity metrics, which struggle with feature ambiguity, MaskField effectively leverages SAM’s object boundaries to double performance. While alternative methods like Gaussian Grouping and Langsplat incorporate SAM for shape consistency, they lack flexibility in scale handling due to their reliance on a single, full segmentation map as supervision. In contrast, MaskField’s support for overlapping masks across multiple scales offers greater adaptability, addressing the scale variation challenge more robustly. Moreover, MaskField naturally disentangles object masks by independently modeling objects through separate query and feature tokens, making object querying straightforward and inherently well-suited to the task. This structure minimizes boundary ambiguity, a common issue with CLIP features, ensuring that objects are distinctly represented and less prone to confusion at the edges.

\subsection{Open-vocabulary Segmentation.}
The 3D-OVS dataset and the Replica dataset are used to evaluate the effectiveness of MaskField in open-vocabulary segmentation. The 3D-OVS dataset provides a diverse collection of long-tail objects in various poses and backgrounds, making it ideal for assessing our method’s performance across complex and varied scenarios. In comparison, the Replica dataset presents more intricate geometries than 3D-OVS. Following established protocols from previous works \cite{kobayashi2022decomposing,zhou2023feature}, we filter out images with unrealistic renderings and defective annotations for both training and evaluation. Additionally, we semi-automatically refine the dataset’s label set using strategies described in DFF \cite{kobayashi2022decomposing} and Feature GS \cite{zhou2023feature} (Appendix D in their respective papers).

\nbf{3D-OVS} We have selected four NeRF-based methods for our analysis: FFD \cite{kobayashi2022decomposing}, Semantic-NeRF \cite{zhi2021place}, LERF \cite{lerf2023}, and 3D-OVS \cite{liu2023weakly}. Additionally, we include results from independently segmenting each test view using 2D open-vocabulary segmentation methods such as LSeg \cite{li2021language}, and OSIDE \cite{xu2023open}. Furthermore, we implement MaskField with 3DGS and compare it with other 3DGS-based methods like Feature 3DGS \cite{zhou2023feature}, Langsplat \cite{qin2023langsplat}, and Gaussian Grouping \cite{ye2023gaussian}. \cref{tab:comparision_iou} details the performance of MaskField alongside baseline methods. The qualitative results are given in \cref{fig:3dovs}. 

MaskField outperforms previous neural field-based methods by a notable margin. Notably, methods like LSeg, ODISE, FFD, and Feature GS relay a fine-tuned CLIP model for boundary precision. However, fine-tuning can damage the extensive open-vocabulary knowledge inherent in CLIP, posing challenges in recognizing long-tailed classes. Although 3D-OVS and LERF have implemented strategies to mitigate the ambiguity in CLIP features, their outcomes are generally less impressive, often marred by overly smoothed features. Langsplat advocates the use of SAM in feature extraction; it requires training three different models in different scales, thus hindering efficiency. By addressing these challenges, MaskField offers a more robust and efficient approach to open-vocabulary 3D segmentation.

\nbf{Replica} The Replica dataset, with its complex geometry, provides an ideal setting to evaluate the effectiveness of feature distillation in challenging 3D structures. To assess MaskField’s performance under these conditions, we follow the protocols established in previous works \cite{kobayashi2022decomposing,zhou2023feature} and compare MaskField with four commonly used feature distillation methods: FFD \cite{kobayashi2022decomposing}, Feature GS \cite{zhou2023feature}, 3D-OVS \cite{liu2023weakly}, and LERF \cite{lerf2023}. For consistency, all methods distill the same features from MaskCLIP \cite{zhou2022extract} and use FeatUP \cite{fu2024featup} for upsampling. The quantitative and qualitative results, presented in \cref{tab:comparision_iou} and \cref{fig:replica}, underscore MaskField’s advantages in handling complex geometries, with SAM segmented boundaries playing a crucial role. By introducing these boundaries, MaskField achieves more precise localization of features, effectively capturing fine-grained structural details in the intricate 3D environments of the Replica dataset. Furthermore, the experiment confirms the effectiveness of the MaskField pipeline, demonstrating its stability in complex geometries. The findings validate mask distillation as a promising alternative to per-pixel distillation in 3D segmentation, showcasing its potential to simplify the distillation process while maintaining high accuracy in challenging 3D settings.

\subsection{Efficiency}
\begin{table}[t]
\centering
\caption{Training efficiency analysis on the sofa scene in 3D-OVS dataset.}
\label{tab:effenciency}
\resizebox{0.95\linewidth}{!}{
\begin{tabular}{@{}c@{\,\,}|ccccc@{}}
\midrule
\midrule
& \multirow{1}{*}{\textbf{Feature}} & \multirow{2}{*}{\textbf{mIoU}} & \multirow{2}{*}{\textbf{Accuracy}} & \multirow{1}{*}{\textbf{Training}} & \multirow{2}{*}{\textbf{Inference}}\\
& \textbf{dimension} & & & \textbf{time} & \\
\midrule
\midrule
LERF\cite{lerf2023} & 512& 27.0  & 43.8 & 19.4 min & 121.4 s\\
3D-OVS\cite{liu2023weakly} & 512& 74.0  & 92.3 & 78 min & 6.6 s\\
Langsplat\cite{qin2023langsplat}& 3& 82.3 & 97.9 & 66 min & 401.9 s\\
Feature GS\cite{zhou2023feature} & 128& 06.7  & 12.4 & 87 min & 6.0 s \\
\midrule
MaskField (NeRF)& 16 & 93.8  & 98.3 & \textbf{5 min} & 7.7 s \\
MaskField (3DGS)& 16& \textbf{94.6} & \textbf{98.7} & 22 min  & \textbf{5.9 s} \\
\midrule
\midrule
\end{tabular}
}
\vspace{-0.6cm}
\end{table}
We analyze the training and inference times of MaskField compared to previous state-of-the-art methods, as shown in \cref{tab:effenciency}. All experiments were conducted over 15,000 iterations, with NeRF-based methods using a batch size of 4096 rays. To ensure consistency, we recalculated total iterations for MaskField, which requires rendering entire images during training, to match the 4096-ray batch size. MaskField achieves notable improvements, efficiently addressing challenges that existing methods face. For instance, while LERF struggles with multi-scale CLIP feature pyramids and 3D-OVS depend on complex regularization, MaskField follows a more streamlined approach. Similarly, MaskField avoids the need for Langsplat’s separate models for different scales and bypasses the slow rasterization that hinders Feature GS due to high-dimensional feature Gaussians. Furthermore, MaskField naturally supports low feature dimensionality without requiring an autoencoder for dimensionality reduction, making inference significantly faster than Langsplat.

In summary, the efficiency of MaskField is driven by two main factors. \textit{Shape and Semantic Distillation Decoupling}: By isolating shape boundaries from semantic information, MaskField simplifies the feature learning process, allowing the model to focus separately on structural and contextual information. This separation enables a more efficient and manageable training procedure compared to multi-scale or joint feature learning. (2) \textit{Avoidance of High-Dimensional Feature Distillation in 3D Space}: MaskField bypasses the intensive task of high-dimensional feature distillation in 3D by employing binary masks, which also naturally solves the ambiguity in CLIP features.

\subsection{Ablation and Analysis}
\begin{table}[t]
  \centering
  \caption{Results with reduced feature dimensions}
  \resizebox{0.95\linewidth}{!}{
    \begin{tabular}{c|cc|cc}
    \toprule
    \toprule
    Scene & \multicolumn{2}{c|}{room0} & \multicolumn{2}{c}{room1} \\
    \midrule
     feature dim & MaskField (3DGS) & Feature GS & MaskField (3DGS) & Feature GS \\
    \midrule
    4     & 74.8& 1.3  & 51.3  & 17.2  \\
    8     & 79.6& 62.3  & 64.2  & 36.9  \\
    16    & 80.9& 74.6  & 69.6  & 56.0  \\
    32    & 81.7& 81.0  & 70.9  & 61.6  \\
    64    & 83.1& 83.4  & 71.9  & 70.3  \\
    128   & 84.3& 84.3  & 74.1  & 73.8  \\
    \bottomrule
    \bottomrule
    \end{tabular}%
    }
  \label{tab:featdim}%
\end{table}%

\nbf{Per-pixel vs. Mask Distillation}
\begin{table}[t]
\definecolor{red}{rgb}{1,0.6,0.6}
\definecolor{orange}{rgb}{1,0.8,0.6}
\definecolor{yellow}{rgb}{1,1,0.6}
  \centering
  \caption{Per-pixel vs. Mask Distillation. }
  \resizebox{0.95\linewidth}{!}{
  \begin{tabular}{c|cc|cc|cc}
    \toprule
    \toprule
    Scene & \multicolumn{2}{c|}{sofa} & \multicolumn{2}{c|}{room0} & \multicolumn{2}{c}{figurines} \\
    \midrule
    Metric & \textbf{mIoU} & \textbf{Acc} & \textbf{mIoU} & \textbf{Acc} & \textbf{mIoU} & \textbf{mBIoU} \\
    \midrule
    Per-pixel (coarse CLIP) & 59.0 & 78.8  & 42.5& 59.1& 30.1& 28.6\\
    Per-pixel (SAM s)  & 71.3 & 83.0  & 49.5& 64.7& 34.9& 29.4\\
    Per-pixel (SAM m) & 67.7  & 81.9  & 51.2& 64.1& 53.4& 49.9\\
    Per-pixel (SAM l)  & 75.5  & 82.1  & 52.4& 64.5& 52.1& 49.0\\
    \midrule
    MaskField(NeRF) & 93.8  & 98.3& \textbf{71.9}& \textbf{80.7}& \textbf{79.5}& \textbf{75.8}\\
    MaskField(3DGS) & \textbf{94.6}& \textbf{98.7}& 68.9& 77.7&  74.2&  71.6\\
    \bottomrule
    \bottomrule
    \end{tabular}%
    }
  \label{tab:pvm}%
  \vspace{-0.6cm}
\end{table}%
In \cref{tab:pvm}, we verify the improvements achieved through mask distillation. Initially, we examine the performance gained by employing SAM-generated masks. Specifically, we perform per-pixel distillation using coarse CLIP features and report its performance. Subsequently, we use masks generated by SAM to compute CLIP embeddings, akin to the approach used in Langsplat \cite{qin2023langsplat}. Each pixel is assigned a CLIP feature corresponding to its associated SAM mask. Given that SAM extracts masks at three different scales, we train three models at three scales and report performance. Direct distillation of per-pixel CLIP features without regularization tends to lead to ambiguity around object boundaries and parts with similar appearances. Using CLIP features with regions segmented by SAM alleviates boundary issues but still suffers from ambiguity related to scale. MaskField naturally incorporates the boundaries delineated by SAM and also integrates different object scales. This is because MaskField allows one pixel to belong to several masks that describe different scales of the object, whereas previous methods with pixel-level distillation only accept one full feature map as supervision. This distinction allows MaskField to manage the scales in the segmentation process more flexibly.

\nbf{Feature Dimension for 3D Gaussian Splatting} 
We explore the impact of feature dimensionality on per-pixel and mask-level distillation, focusing on evaluating the quality of distilled features while reducing the ambiguity inherent in CLIP features. In other words, we are interested in understanding how MaskField performs when provided with ideal dense pixel-aligned vision-language features. To achieve this, we utilize features from LSeg \cite{li2021language}, a fine-tuned vision-language model known for achieving precise object boundaries and accurate segmentation within its domain-specific datasets, such as indoor scenes like Replica. By using LSeg’s predictions as ground truth, we distill its features into both Feature GS and MaskField, experimenting across various feature dimensions. The outcomes are detailed in \cref{tab:featdim}.

Remarkably, MaskField demonstrates high mIoU and accuracy even at lower feature dimensions. This strength arises from MaskField’s distinct approach of separating mask and feature distillation. Each Gaussian feature in MaskField is tailored to capture shape information effectively, facilitating dimensionality reduction without compromising quality. This attribute makes MaskField particularly advantageous for 3DGS, supporting efficient feature distillation with reduced dimensionality and minimizing computational costs while maintaining segmentation accuracy.


\section{Conclusion}
\label{sec:conclusion}
In this paper, we present MaskField, a novel learning paradigm for 3D segmentation using neural fields. Our primary goal is to demonstrate that mask distillation can serve as an effective alternative to per-pixel distillation in 3D segmentation. Mask distillation aligns naturally with neural fields by sidestepping the complexities of managing pixel-aligned, high-dimensional CLIP features, thereby improving efficiency. Importantly, for techniques like 3D Gaussian Splatting (3DGS), MaskField naturally eliminates the need for dimensionality reduction strategies. Our extensive experiments show that MaskField not only outperforms previous state-of-the-art methods but also achieves remarkable efficiency, surpassing prior techniques with just 5 minutes of training. We hope that MaskField provides valuable insights and encourages further research in this promising field.

{\small
\bibliographystyle{ieeenat_fullname}
\bibliography{11_references}
}

\ifarxiv \clearpage \appendix In this supplement, we first conduct more experiments related to the proposed MaskField to discuss the motivation and limitations. We also include more qualitative results in different scenes. Finally, we will add more details about the experimental settings and implementations.

\section{More Experiments.}

\nbf{Effects of Limited Input}
\begin{table}[t]
  \centering
  \caption{{Quantitative comparisons in 3 selective scenes with limitted input.}}
  \resizebox{1.0\linewidth}{!}{
  \begin{tabular}{c|c|cc|cc|cc}
    \toprule
    \toprule
    & Scene & \multicolumn{2}{c|}{sofa} & \multicolumn{2}{c|}{room0} & \multicolumn{2}{c}{figurines} \\
    \midrule
    & Metric & \textbf{mIoU} & \textbf{Acc} & \textbf{mIoU} & \textbf{Acc} & \textbf{mIoU} & \textbf{mBIoU} \\
    \midrule
    \multirow{2}{*}{Full View}
   & \textbf{ NeRF-based } &93.8  &98.3 &71.9 &80.7 &79.5 &75.8 \\
   & \textbf{ 3DGS-based } &94.6  &98.7 &68.9 &77.7 &74.2 &71.6 \\
    \midrule
    \multirow{2}{*}{50\% View}
   & \textbf{ NeRF-based } &93.0&98.1&69.0&77.6&78.8&74.1\\
   & \textbf{ 3DGS-based } &92.5 &98.1 &67.1 &74.9 &72.3 &69.5 \\
    \midrule
    \multirow{2}{*}{10\% View}
   & \textbf{ NeRF-based } &91.5&98.0&66.9&74.1&70.0&64.6\\
   & \textbf{ 3DGS-based } &78.0 &89.7 &66.5 &74.3 &69.8 &64.6 \\
    \bottomrule
    \bottomrule
    \end{tabular}%
    }
  \label{tab:limited_replica}%
\end{table}%
In this experiment, we evaluate the performance of MaskField when trained using a limited subset of images for mask and feature extraction, while utilizing all available images during the reconstruction phase. This setup simulates scenarios where only sparse labeled data or limited computational resources are available for training. By comparing segmentation performance under these constraints, we aim to assess MaskField’s ability to generalize and maintain multi-view consistency despite restricted input during training. The results are present in \cref{tab:limited_replica}.

The NeRF-based method consistently achieves higher metrics in settings with reduced views, particularly evident in the 10\% view scenario. This superiority can be attributed to the continuous nature of NeRF’s MLP representation, which inherently provides better locality and smoothness, even with sparse inputs. In contrast, the 3DGS-based approach relies on independent features for each Gaussian splat. Under limited viewpoints, many Gaussians receive insufficient coverage or are entirely unobserved, leading to suboptimal optimization. This limitation is most pronounced in the “sofa” scene, where the mIoU of the 3DGS-based method drops significantly with only 10\% view coverage. While 3DGS-based methods show competitive performance under full-view conditions due to their efficient representation, their reliance on direct feature learning makes them less robust in highly sparse settings.

These results highlight the trade-offs between NeRF-based and 3DGS-based methods, with the former excelling in locality and the latter offering computational advantages in rendering.

\section{Additional Visualizations.}
\nbf{Learned Masks}
\begin{figure*}
    \centering
    \includegraphics[width=\linewidth]{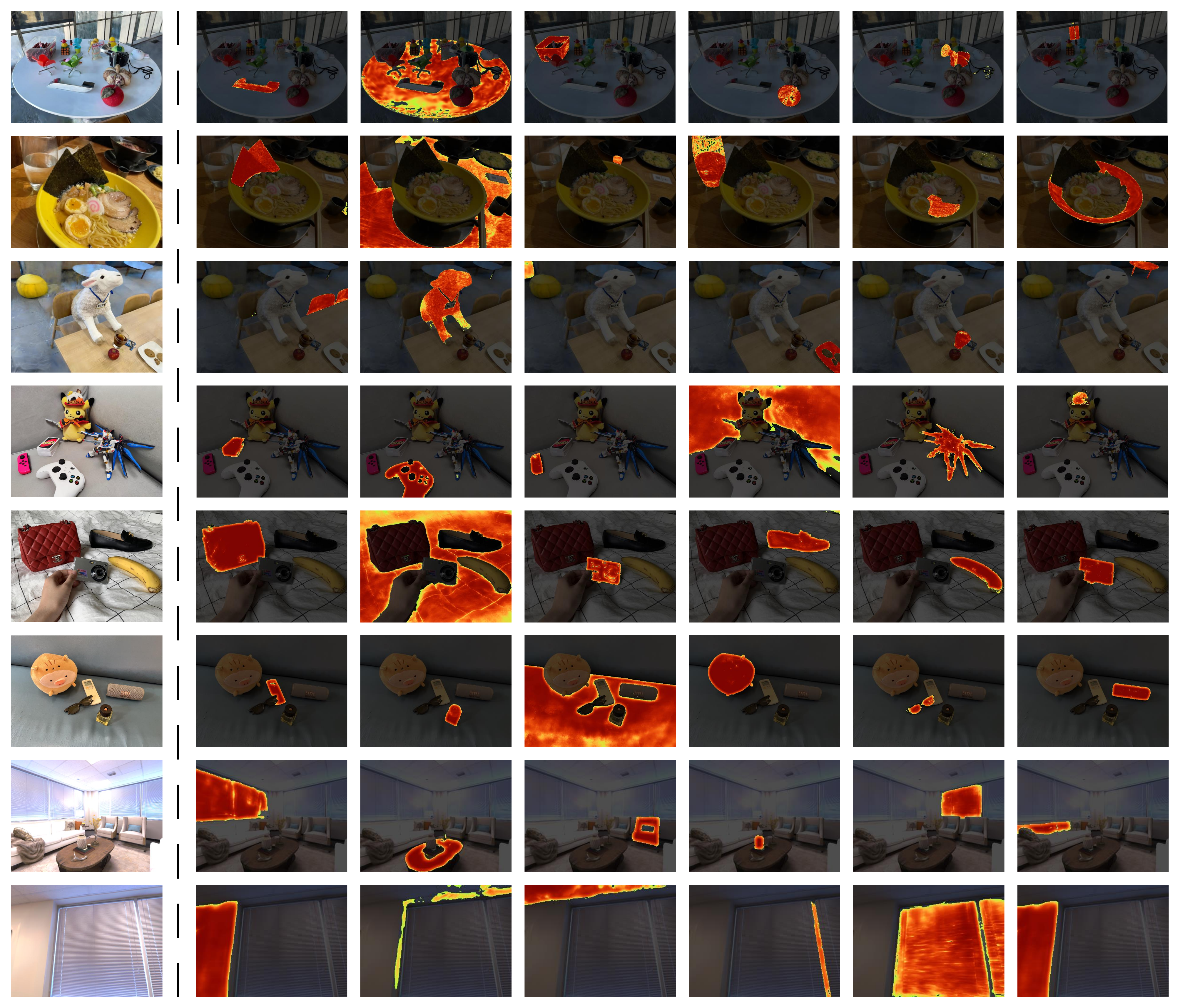}
    \caption{\textbf{Qualitative results of distilled mask.} \textbf{Figure 1. Qualitative results of distilled masks.} Visualization of masks distilled into the mask neural field. Each row represents a scene, and the columns show the input image, and the masks distilled by our method. Our approach naturally and effectively captures object boundaries segmented by SAM.}
    \label{fig:mask_supp}
\end{figure*}
The qualitative results in \cref{fig:mask_supp} illustrate the ability of MaskField to effectively distill SAM-generated masks into the mask neural field. The distilled masks exhibit precise object boundaries, demonstrating the method’s capability to preserve spatial details during the distillation process. Across diverse scenarios, including tabletop and indoor scenes, the method achieves consistent and robust segmentation, highlighting its adaptability to varied environments. These results emphasize the effectiveness of MaskField in overcoming ambiguities in CLIP features, achieving both high boundary accuracy and semantic fidelity.

\nbf{More Visualization on Different Scenes}
\begin{figure*}
    \centering
    \includegraphics[width=\linewidth]{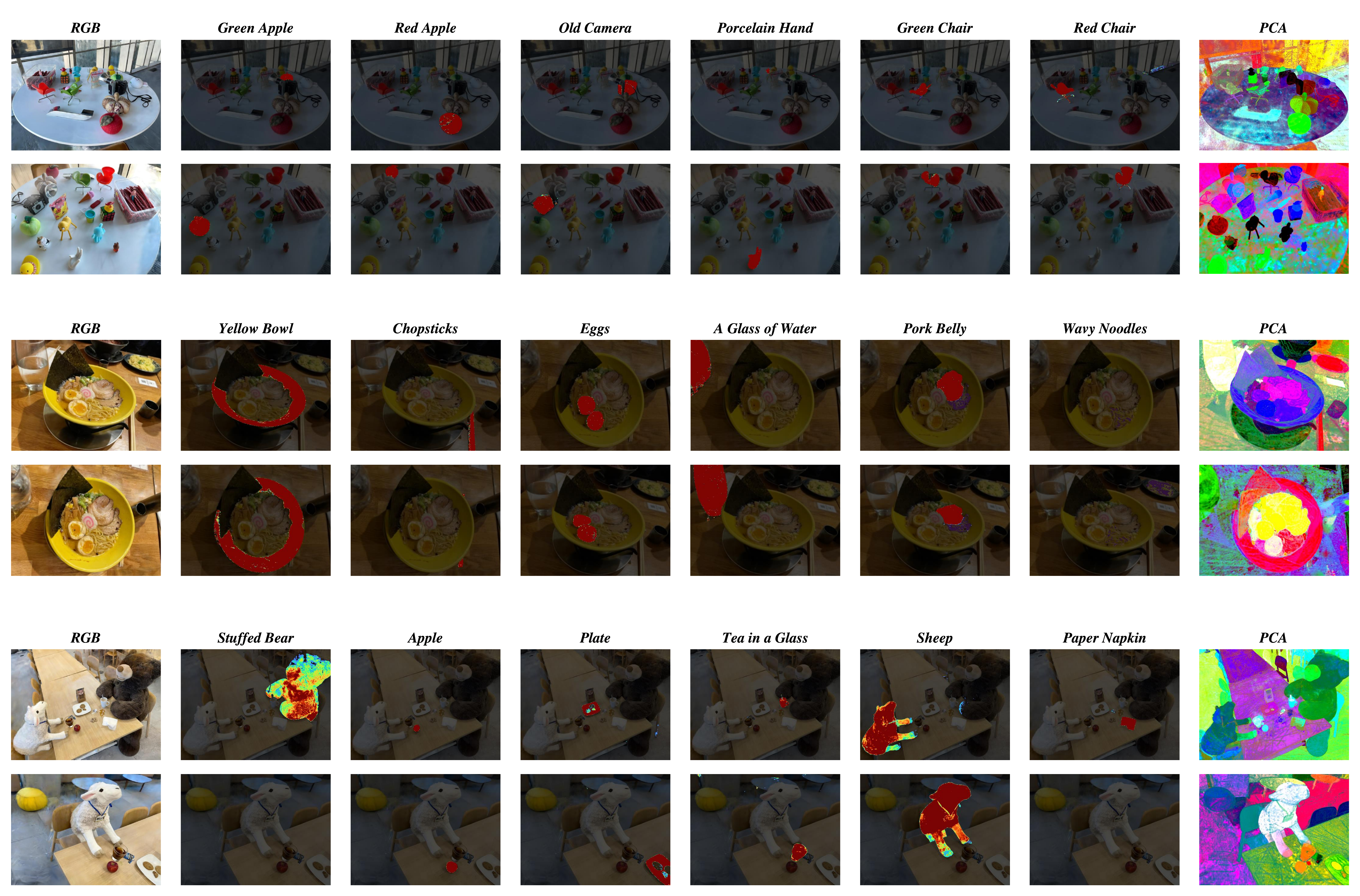}
    \caption{\textbf{Qualitative comparisons of 3 different scenes in LERF-Mask dataset.} Our method successfully gives accurate segmentation maps.}
    \label{fig:lerf_supp}
\end{figure*}
\begin{figure*}
    \centering
    \includegraphics[width=\linewidth]{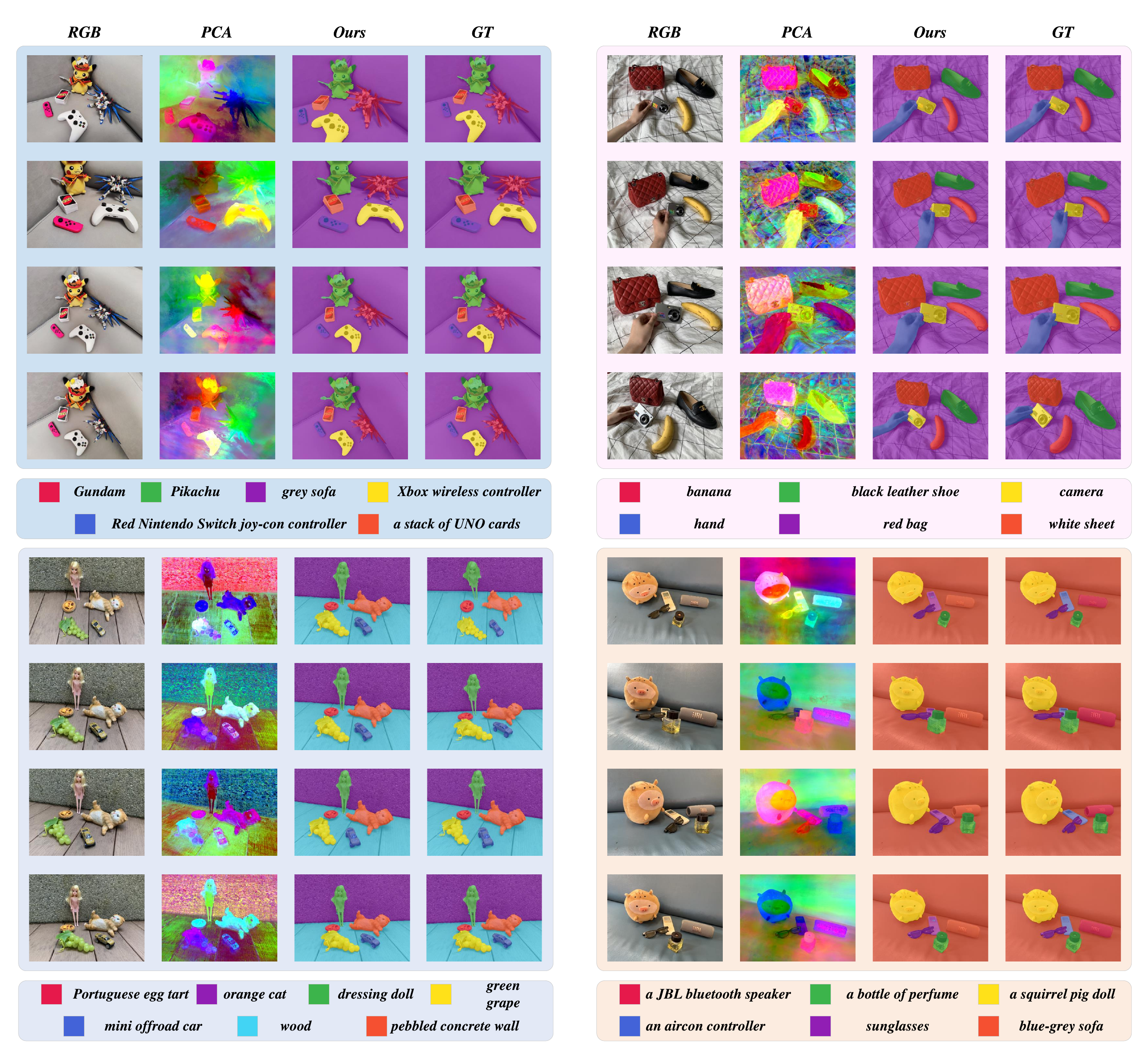}
    \caption{\textbf{Qualitative comparisons of 4 different scenes in 3DOVS dataset.} Our method successfully recognizes long-tailed objects and gives accurate segmentation maps.}
    \label{fig:3dovs_supp}
\end{figure*}
\begin{figure*}
    \centering
    \includegraphics[width=\linewidth]{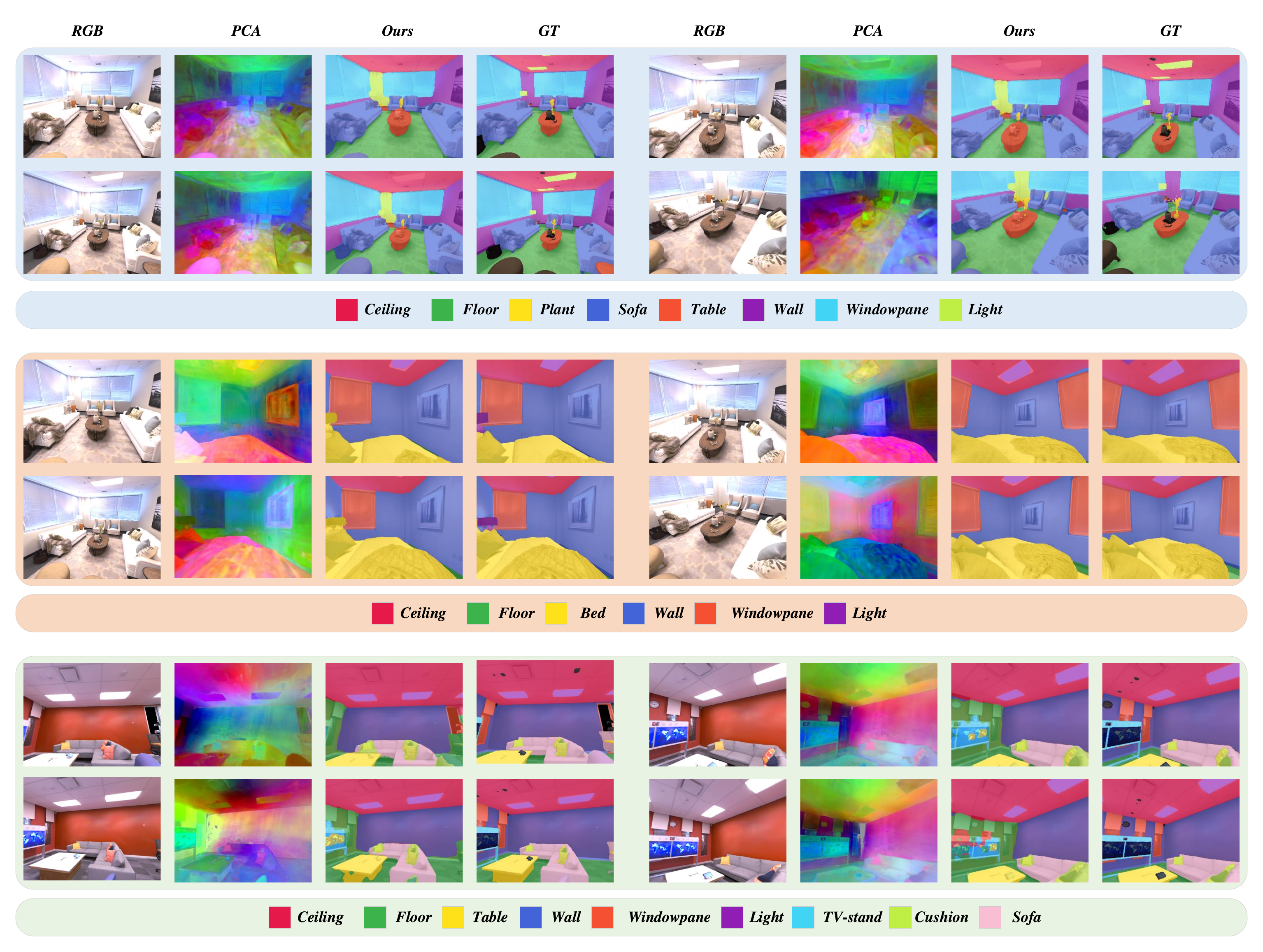}
    \caption{\textbf{Qualitative comparisons of 3 different scenes in Replica dataset.} Our method successfully gives accurate segmentation maps.}
    \label{fig:replica_supp}
\end{figure*}
We further visualize results on the LERF-Mask, 3D-OVS, and Replica datasets to evaluate the performance of MaskField across diverse scenarios. The results are shown in \cref{fig:lerf_supp}, \cref{fig:3dovs_supp} and \cref{fig:replica_supp}. The proposed MaskField consistently delivers accurate segmentation results while maintaining multi-view consistency. Notably, in cases where certain objects are not visible in the given viewpoints, MaskField correctly produces blank retrieval maps, demonstrating its robustness and ability to avoid false positives. This indicates that MaskField effectively incorporates geometric cues and semantic priors, ensuring reliable predictions even for occluded or unobserved regions. These results highlight MaskField’s strength in achieving precise segmentation while respecting the inherent limitations of the available viewpoints.

\section{More Implementation Details.}
\nbf{Datasets and Metrics} 
To validate the effectiveness of MaskField, we conduct experiments on three benchmark datasets, each chosen for its unique challenges and suitability for evaluating 3D open-vocabulary segmentation methods.
The 3D-OVS dataset comprises diverse scenes with long-tailed object categories, making it ideal for testing the open-vocabulary capabilities of segmentation models. The dataset presents objects in varied poses and environments, challenging models to generalize across complex scenarios. Following prior works \cite{liu2023weakly, qin2023langsplat}, we evaluate on all 10 scenes to ensure a comprehensive assessment. Unlike methods that rely on subsets, our use of the entire dataset provides a more robust and unbiased comparison. Replica offers intricate indoor scenes with fine-grained geometric details, making it an excellent choice for evaluating segmentation in complex 3D environments. The dataset emphasizes precise feature localization and boundary delineation, which are critical for challenging segmentation tasks. LERF-Mask provides a diverse set of scenes designed to test open-vocabulary query. The dataset’s inclusion of varied object scales and textures is well-suited for analyzing MaskField’s ability to handle ambiguous features.

For quantitative evaluation, we follow established protocols \cite{liu2023weakly, qin2023langsplat, ye2023gaussian} and use mean Intersection-over-Union (mIoU) and mean Boundary IoU (mBIoU) to evaluate open-vocabulary querying, while accuracy (Acc) is used for open-vocabulary segmentation. These metrics comprehensively measure the model’s effectiveness and overall performance across different datasets.

Our selection of datasets ensures that MaskField is evaluated under diverse conditions, highlighting its robustness and generalizability in handling open-vocabulary segmentation across varied 3D environments.

\nbf{Implementation of MaskField} 
We developed two versions of MaskField: MaskField (NeRF) and MaskField (3DGS). For MaskField (NeRF), our implementation builds upon the 3D-OVS \cite{liu2023weakly} codebase, adhering strictly to the original settings and hyperparameters to ensure a fair comparison. Likewise, for MaskField (3DGS), we based our implementation on the methods described in Feature GS \cite{zhou2023feature}, carefully aligning the hyperparameters to facilitate consistent and comparable results across different models. This allows us to accurately evaluate the performance improvements and efficiencies that MaskField introduces within various 3D segmentation frameworks. For bipartite matching of masks, we utilized the implementation available in \textit{scipy}\footnote{https://pypi.org/project/scipy/}. Similar to previous works \cite{qin2023langsplat}, we employ a mean convolution filter with a size of 10 to smooth the mask maps.

\nbf{Open Vocabulary Query in LERF-Mask Dataset} 
Given a text query, we compute relevance scores following LERF \cite{lerf2023}, using predefined canonical phrases such as \textit{"object"}, \textit{"things"}, \textit{"stuff"}, and \textit{"texture"}. For each query, we obtain \(N_{K}\) relevance scores and their corresponding masks. These scores are normalized, and masks are filtered based on a relevance threshold of 0.9. Additionally, masks with low activation are removed. The remaining masks are then aggregated to produce the final prediction. For the comparison experiment, we reproduce experiments established in \cite{ye2023gaussian} and find the difference is minimal, so we directly borrow their results.

\nbf{Open Vocabulary Segmentation in 3D-OVS Dataset}
Unlike prior works \cite{liu2023weakly, qin2023langsplat} that evaluate on a subset of the 3D-OVS dataset, we report performance on all 10 scenes to provide a more comprehensive evaluation. For comparisons, we identify LERF \cite{lerf2023}, Langsplat \cite{qin2023langsplat}, and 3D-OVS \cite{liu2023weakly} as the most relevant baselines, as these methods train feature fields directly from original CLIP features. Additionally, we consider segmentation methods such as FFD \cite{zhi2021place}, Feature GS \cite{zhou2023feature}, and Gaussian Grouping \cite{ye2023gaussian}, which leverage fine object boundaries from fine-tuned open-vocabulary segmentation models or utilize the open-vocabulary capabilities of pre-trained detectors. However, this reliance on fine-tuned models can be a double-edged sword: while they perform well within their trained domains, they often struggle to recognize long-tailed or uncommon objects. We implement these baseline methods strictly follow their open-source code base and hyperparamters.

\nbf{Open Vocabulary Segmentation in Replica Dataset}
Following the dataset selection criteria in \cite{zhi2021place, kobayashi2022decomposing, zhou2023feature}, each scene consists of 80 images captured along a randomly chosen trajectory, with every 10th image selected for testing. Consistent with the protocols established in \cite{kobayashi2022decomposing, zhou2023feature}, we manually merge semantically similar categories, such as "rugs" and "floor," and exclude categories that cannot be effectively segmented due to the coarse granularity of CLIP features.

For the comparative experiments, we evaluate the effectiveness of feature distillation by comparing MaskField with four commonly used feature distillation methods, all based on MaskCLIP features. In the feature dimensionality analysis experiments, our goal is to assess MaskField’s ability while minimizing the inherent advantage of SAM’s shape priors. To this end, we use features from LSeg \cite{li2021language}, a fine-tuned vision-language model that provides precise object boundaries and accurate segmentation within domain-specific datasets, such as indoor scenes like Replica. For this experiment, we utilize text labels from the ADE20K dataset.

\section{Limitations and Future Works}
While MaskField demonstrates strong performance across various datasets, its effectiveness depends on the reliability of SAM-generated masks. Although SAM generally produces accurate and consistent masks, there are instances where its predictions may be less reliable. In such cases, the matching process between SAM masks and MaskField queries can become unstable, potentially affecting overall segmentation quality. To address this, future work could explore methods to improve the robustness of the matching process, such as incorporating multi-view prompts for SAM to enhance mask consistency or developing techniques to adaptively refine unreliable masks during training.

Additionally, MaskField’s current implementation is limited to static scenes due to the constraints of existing neural field methods. Extending MaskField to dynamic scenes introduces new challenges, such as handling temporal consistency and adapting to changing object geometries. Future research could investigate how MaskField can be adapted to dynamic environments, potentially by integrating advances in neural dynamic fields or leveraging temporal information from multi-view sequences.

By addressing these limitations, MaskField could further enhance its applicability to a wider range of real-world scenarios, including dynamic and highly complex environments.

 \fi

\end{document}